\documentclass[10pt,twocolumn,letterpaper]{article}

\usepackage[pagenumbers]{wacv} 

\usepackage{graphicx}
\usepackage{amsmath}
\usepackage{amssymb}
\usepackage{booktabs}
\usepackage[accsupp]{axessibility}  
\usepackage[flushleft]{threeparttable}
\usepackage{amssymb} 
\usepackage{multirow} 
\usepackage{multicol} 
\usepackage[shortcuts,acronym]{glossaries}
\newacronym{cnn}{CNN}{Convolutional Neural Network
}
\newacronym{dgcnn}{DGCNN}{Dynamic Graph Convolutional Neural Network}

\newacronym{knn}{kNN}{k-Nearest-Neighbor}
\newacronym{gnn}{GNN}{Graph Neural Network}
\newacronym{mlp}{MLP}{Multi-Layer Perceptron}
\newacronym{rcs}{RCS}{Radar Cross Section}
\newacronym{ascb}{ASCB}{Adaptive Sparse Convolution Block}
\usepackage{pifont}
\newcommand{\cmark}{\ding{51}}%
\newcommand{\xmark}{\ding{55}}%

\usepackage[pagebackref,breaklinks,colorlinks]{hyperref}

\usepackage[capitalize]{cleveref}
\crefname{section}{Sec.}{Secs.}
\Crefname{section}{Section}{Sections}
\Crefname{table}{Table}{Tables}
\crefname{table}{Tab.}{Tabs.}

\def\wacvPaperID{747} 
\def\confName{WACV}
\def\confYear{2025}

\begin{document}

\title{GET-UP: GEomeTric-aware Depth Estimation with Radar Points UPsampling}


\author{
    Huawei Sun$^{1,2}$, Zixu Wang$^{1,2}$, Hao Feng$^{1}$,  Julius Ott$^{1,2}$,  Lorenzo Servadei$^{1}$ , Robert Wille$^{1}$\\
    \\
    $^{1}$ Technical University of Munich, Munich, Germany \\
    $^{2}$ Infineon Technologies AG, Neubiberg, Germany
}

\maketitle

\begin{abstract}
Depth estimation plays a pivotal role in autonomous driving, facilitating a comprehensive understanding of the vehicle's 3D surroundings. Radar, with its robustness to adverse weather conditions and capability to measure distances, has drawn significant interest for radar-camera depth estimation. However, existing algorithms process the inherently noisy and sparse radar data by projecting 3D points onto the image plane for pixel-level feature extraction, overlooking the valuable geometric information contained within the radar point cloud. To address this gap, we propose GET-UP, leveraging attention-enhanced Graph Neural Networks (GNN) to exchange and aggregate both 2D and 3D information from radar data. This approach effectively enriches the feature representation by incorporating spatial relationships compared to traditional methods that rely only on 2D feature extraction. Furthermore, we incorporate a point cloud upsampling task to densify the radar point cloud, rectify point positions, and derive additional 3D features under the guidance of lidar data. Finally, we fuse radar and camera features during the decoding phase for depth estimation. We benchmark our proposed GET-UP on the nuScenes dataset, achieving state-of-the-art performance with a $15.3\%$ and $14.7\%$ improvement in MAE and RMSE over the previously best-performing model. Code: \url{https://github.com/harborsarah/GET-UP}
\end{abstract}

\vspace{-6mm}

\section{Introduction}
Understanding the 3D environment surrounding the ego vehicle is essential in the autonomous driving field,  requiring the estimation of dense depth maps for 3D scene reconstruction. While learning-based monocular depth estimation methods \cite{adabins,binsformer,bts,cspn,dorn,eigen2014depth,p3depth} have outperformed traditional monocular-based approaches \cite{trad1,trad2,trad3} in accuracy, they are still constrained by the lack of robust geometric constraints. To address this limitation, methods~\cite{lidar_2d1,lidar_2d2,lidar_2d3,lidar_2d4,lidar_2d5,lidar_2d6,lidar_2d7,lidar_2d8} leveraging both depth sensors (i.e. LiDAR) and RGB images to first project LiDAR points onto the image plane, resulting in a sparse depth map. However, these methods require additional tasks such as surface normal estimation for improved feature learning \cite{lidar_2d8}. 

\begin{figure}[ht]
\centering
\vspace{-3mm}
\includegraphics[width=0.98\linewidth]{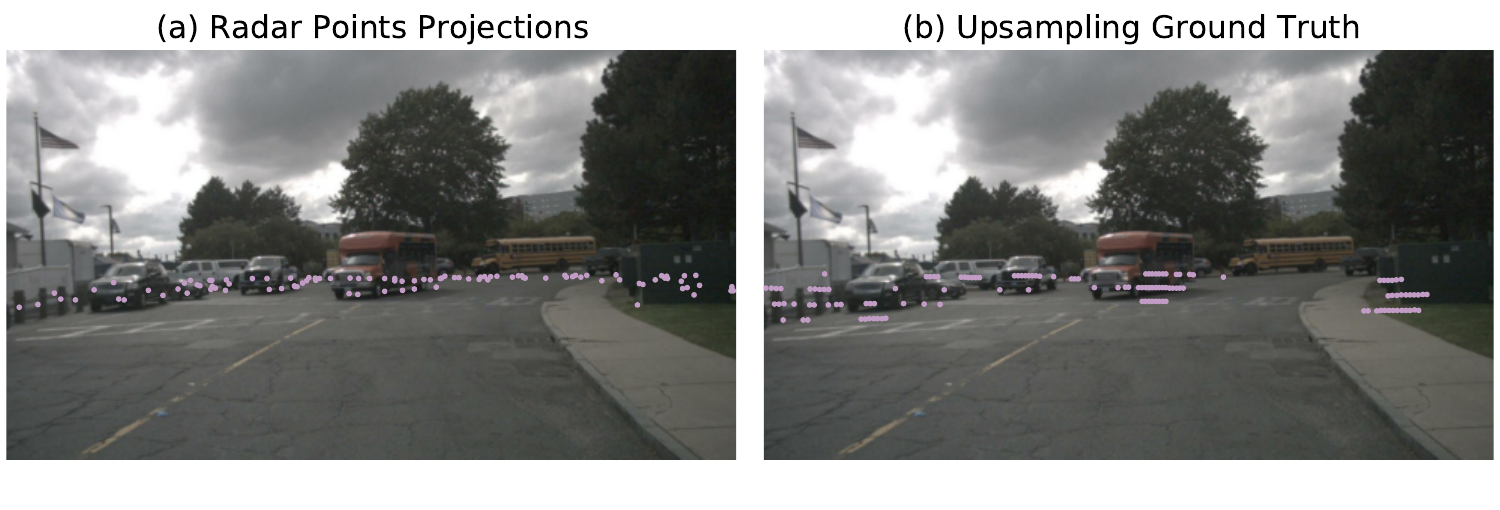}

\vspace{-4mm}
   \caption{Visualization of projected radar points compared with the selected LiDAR points employed for point cloud upsampling.}
\label{fig:upsampling_vis}
\vspace{-4mm}

\end{figure}

\begin{figure}[t]
\centering
\vspace{-2mm}
\includegraphics[width=0.99\linewidth]{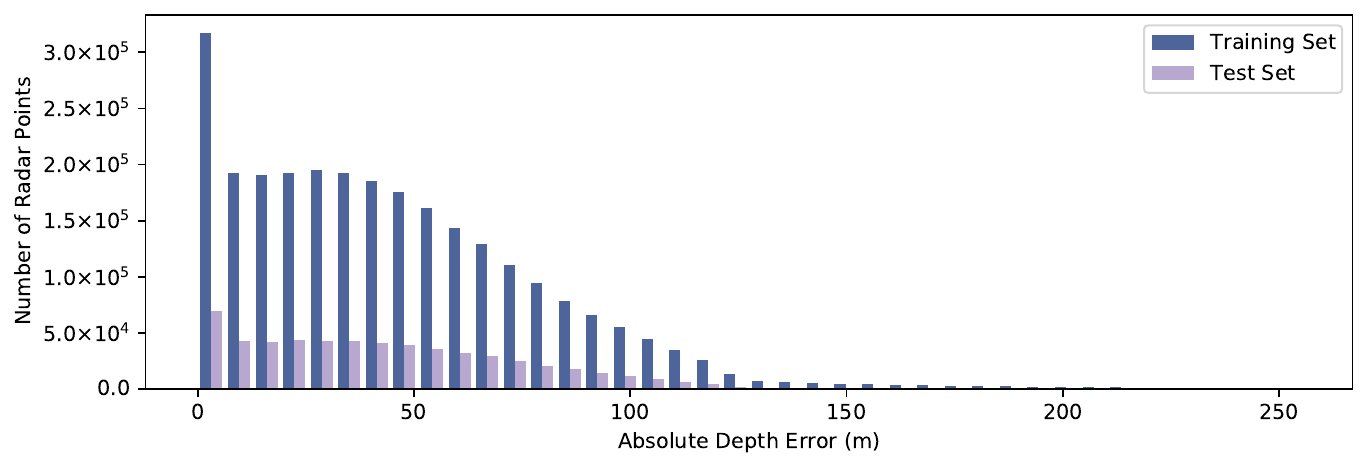}
\vspace{-3mm}
   \caption{Absolute depth difference between each radar point and its corresponding nearest LiDAR point.}
\label{fig:dist_err}
\vspace{-6mm}
\end{figure}

Although LiDAR provides detailed information about 3D scenes, it is prohibitively expensive and sensitive to weather conditions.
In contrast, radar offers robust performance in all weather conditions and is more cost-effective than LiDAR.
However, the absence of height information in radar data and noisy characteristics lead to significant errors when projecting radar points onto the image plane. To illustrate how this discrepancy complicates depth estimation, we analyzed the absolute depth differences between each radar point and its nearest corresponding LiDAR point on the 2D image plane across the dataset.
As shown in Figure \ref{fig:dist_err}, the depth values associated with radar points frequently deviate significantly from those of LiDAR, which serves as the ground truth. This discrepancy highlights why LiDAR-camera depth completion algorithms, which typically propagate depth information from LiDAR points to surrounding pixels, are ill-suited for radar-camera setups, indicating that radar-specific algorithms need to be developed.

Studies like~\cite{radar_proj,radarproj2} directly project radar points onto the image plane, resulting in sparse and ambiguous radar projection maps. Others extend the height of each radar point~\cite{crf,mcafnet,dorn_radar,enhance} or adopt two-stage processes producing semi-dense radar depth maps \cite{radarnet,rc-pda,sun2024cafnet} to mitigate this issue. However, these methods often distort 3D geometric details, thereby limiting feature extraction in 2D space and introducing further noises into the radar data by directly altering the radar input. 





To address these challenges, we propose GET-UP, a novel radar-camera depth estimation framework that utilizes radar input across two domains. Firstly, the 3D radar points are projected onto the image plane and densified by proposed \acrfull{ascb}, followed by ResNet-18~\cite{resnet}, yielding 2D radar features. 
Secondly, we employ an attention-enhanced \acrfull{dgcnn} to capture 3D information and facilitate interaction with 2D features.

A key contribution of our method is the incorporation of point cloud upsampling as an auxiliary task. This block refines the radar features from precise LiDAR data, addressing the inherent ambiguity problem in radar point positions and ensuring that the densification process does not introduce extraneous noise into the radar inputs. Since the depth ground truth is generated from LiDAR data, no additional resource is needed for this training process. Figure \ref{fig:upsampling_vis} visualizes the radar and selected LiDAR points projections, serving as the ground truth for the upsampling task.
To the best of our knowledge, our work is the first radar-camera depth estimation method to explicitly consider the 3D geometric information in radar point clouds. Moreover, this study pioneers using a point cloud upsampling strategy to effectively address the challenge of radar data sparsity. In summary, our principal contributions are:
\vspace{-2mm}
\begin{itemize}
    \item A novel depth estimation framework is proposed that uniquely takes advantage of 2D and 3D representations of radar data.
\vspace{-2.5mm}
    \item An attention-enhanced DGCNN model is designed to adeptly extract 3D features while preserving the integrity of 2D spatial information.
\vspace{-2.5mm}
    \item We present two innovative and effective approaches to address radar data sparsity: the \acrshort{ascb} for improving feature extraction in the 2D space and a dedicated point cloud upsampling task for enriched radar point representation from the 3D perspective.
\vspace{-2.5mm}
    \item Our GET-UP model outperforms existing state-of-the-art radar-camera depth estimation techniques on the nuScenes dataset \cite{nuscenes}.
\end{itemize}

\label{sec:intro}
\vspace{-4mm}
\section{Related Work}
\label{sec:related_work}
\vspace{-1mm}
\paragraph{Geometry-aware Depth Completion.}
Initial studies in LiDAR-camera depth completion \cite{lidar_2d1,lidar_2d2,lidar_2d3, lidar_2d4,lidar_2d5,lidar_2d6} predominantly perform depth completion within the 2D image plane by projecting sparse LiDAR points onto it, which fall short of capturing the underlying 3D geometric information. Instead, the following studies also extract features from the 3D perspective. Xiong \etal \cite{s2d_GNN} employ a \acrshort{gnn} by treating each image pixel as a graph node and establishing connections based on the \acrfull{knn} principle in 3D space. Further advancements include graph propagation techniques as seen in \cite{adaptive_context}, enhancing multi-modal feature integration. Moreover, Point-Fusion\cite{PointFusion}, FuseNet \cite{fuseNet}, and \cite{PointDC} 
extract 3D features from 3D LiDAR points and consolidate 2D and 3D features. 

Nevertheless, these approaches are constrained by relying on a predefined number of LiDAR points as input, which cannot handle the various number of radar points.
\vspace{-4.5mm}
\paragraph{Radar-Camera Depth Estimation. }
Radar point clouds are significantly sparser and noisier than LiDAR, presenting a challenge for generating dense depth maps from images and radar data. Lin \etal \cite{lin2020depth} directly project radar points onto the image plane, yielding highly-sparse and ambiguous radar maps. To mitigate the sparsity issue, \cite{dorn_radar,li2023sparse} extend radar points vertically, creating denser radar projection maps. 
Differing from direct radar-to-image projection, \cite{rc-pda, radarnet, sun2024cafnet} propose two-stage architectures that explore one-to-many mapping from radar data to image pixels in the first stage, producing denser intermediate radar data for subsequent depth prediction. 

However, the densify processes in the existing studies introduce further noises into the radar data since they directly modify the radar input. Furthermore, the projection process overlooks the 3D geometric information of radar data.
\vspace{-3.5mm}
\paragraph{Point Cloud Upsampling. }
Point cloud upsampling is a typical task for point cloud densification, which is designed to transform sparse and noisy point clouds into denser and cleaner counterparts \cite{pu_cascade,pueva,pugan,pugcn,punet}. This procedure typically begins with extracting point features, followed by point expansion and coordinate reconstruction, a methodology initially introduced by PU-Net \cite{punet}. Subsequently, PU-GAN \cite{pugan} innovated by incorporating an adversarial network to optimize point distribution. Further, \acrshort{gnn}s are utilized in PU-GCN \cite{pugcn} for both feature extraction and expansion phases to improve point cloud quality. 
Du \etal \cite{pu_cascade} advances by introducing a cascaded refinement network that employs a residual learning approach for incremental improvements. Nevertheless, point cloud upsampling task has not been used by existing radar-camera depth estimation methods to mitigate the radar point sparsity problem.

In this study, diverging from conventional radar-camera depth estimation techniques, which distort 3D geometric clues, we advance feature extraction by considering both 2D radar projection maps and 3D radar points. To address the challenges of sparsity and ambiguity inherent in radar point clouds, we are the first to introduce a point cloud upsampling module into the depth estimation task. This module, distinctively utilizing existing LiDAR points as ground truth, aims to both densify the radar data and enhance the precision of radar point positioning.

\begin{figure*}[t]
\centering
\includegraphics[width=0.9\linewidth]{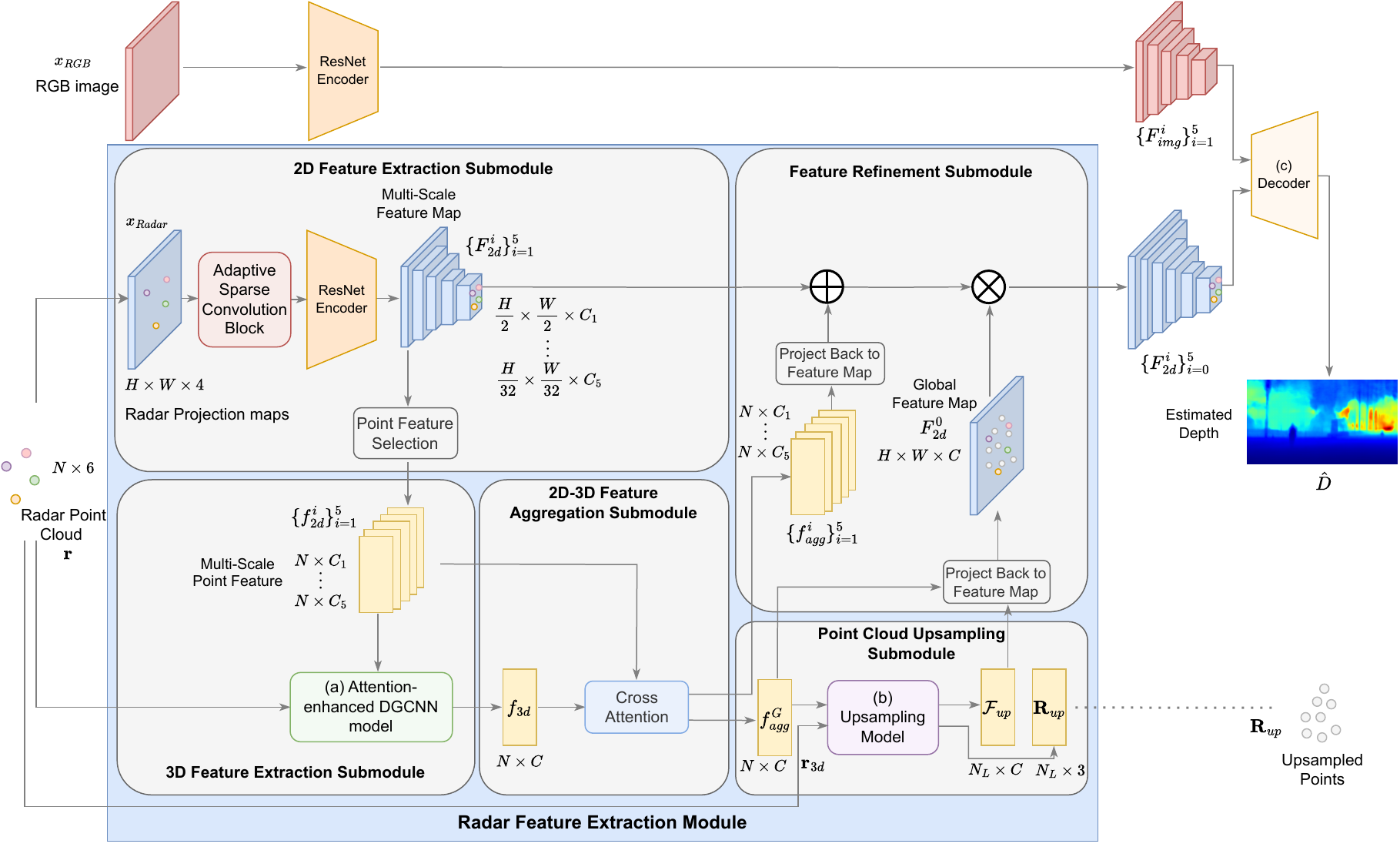}
\vspace{-1mm}

   \caption{Model Architecture: The input image is processed through a ResNet encoder to extract features. Concurrently, radar data are processed by a specially designed radar feature extraction module, comprising five submodules, to yield refined radar features and upsampled points. These radar and image features are then integrated within the decoder to produce the estimated dense depth map. Detailed illustrations of the blocks (a), (b), and (c) are provided in Fig. \ref{fig:DGCNN}, \ref{fig:upsampling}, and \ref{fig:bts_decoder}, respectively.}
\label{fig:model}
\vspace{-4mm}
\end{figure*}

\section{Approach}
\label{sec:approach}
\vspace{-1mm}
This section introduces the innovations of our work. In Sec. \ref{subsec:model}, we explain the model architecture. Subsequently, our proposed radar feature extraction module, including five submodules, is described in Sec. \ref{subsec:radar_feat_ext}. Finally, we present our decoder of the depth estimation task in Sec. \ref{subsec:decoder}.

\subsection{Model Architecture}
\label{subsec:model}

As visualized in Figure \ref{fig:model}, our model processes two key inputs: an image $x_{RGB} \in \mathbb{R}^{H \times W \times 3}$ and a radar point cloud $\textbf{r} \in \mathbb{R}^{N \times C_{r}}$, where $N$ denotes the number of radar points in the current frame and $C_{r}$ is the number of features carried by the radar. Radar projection map $x_{Radar} \in \mathbb{R}^{H \times W \times C_{R}}$ is generated by projecting these $N$ points onto the image plane, carrying specific attributes such as depth, velocities, and \acrfull{rcs}. Additionally, the point cloud $\textbf{r}$ encompasses 3D positional data.

The RGB image $x_{RGB}$ is processed through a ResNet-34 encoder \cite{resnet}, yielding multi-scale features $\{F_{img}^{i}\}_{i=1}^{5}$. Concurrently, $x_{Radar}$ and $\textbf{r}$ are processed by a dedicated radar feature extraction module, designed to extract coherent radar features by aggregating 2D and 3D information with five submodules.
Notably, it includes a point cloud upsampling submodule aimed at leveraging precise LiDAR point positions to adjust and densify radar point representations. This submodule efficiently extracts features reflective of LiDAR positions, which are then used to enhance the radar-derived features. The output of this radar feature extraction process is a set of refined radar features $\{F_{2d}^{i}\}_{i=0}^{5}$, including a generated feature $F_{2d}^{0}$ with the same size as the input image, and an upsampled 3D point cloud $\textbf{R}_{up}$. Sec. \ref{subsec:radar_feat_ext} provides a more detailed explanation of this process.

These processed image and radar features are subsequently fused through a gated fusion mechanism \cite{radarnet} and then fed into a depth estimation model \cite{bts}. This final step produces a comprehensive dense depth map $\hat{D} \in \mathbb{R}^{H \times W}$. More details are available in Sec. \ref{subsec:decoder}.
\subsection{Radar Feature Extraction Module}
\label{subsec:radar_feat_ext}

The radar feature extraction module comprises five distinct submodules with $x_{Radar}$ and $\textbf{r}$ as inputs. Initially, the \emph{2D feature extraction submodule} processes $x_{Radar}$ to generate multi-scale feature maps. Subsequently, these maps, jointly with $\textbf{r}$, are input into the \emph{3D feature extraction submodule}, to distill 3D geometry-aware features. 
Afterwards, the \emph{2D-3D feature aggregation submodule} processed the 2D and 3D features, yielding enhanced and more reliable 3D feature representations.
The obtained 3D features, integrated with $\textbf{r}$, undergo further enhancement in the \emph{point cloud upsampling submodule} to increase data density and precision. Lastly, a \emph{feature refinement submodule} is employed to precisely refine the radar features, leveraging both the 2D spatial and 3D geometric information to obtain a comprehensive feature representation.

\vspace{-2mm}
\subsubsection{2D feature extraction submodule. }
\vspace{-1mm}
We introduce \acrshort{ascb} with sparse convolution layers \cite{sparsecnn} to address the challenge of highly sparse radar projections. This component adaptively adjusts the convolution kernel size based on the depth information of radar points. Initially, $x_{Radar}$ undergoes processing by the \acrshort{ascb}, followed by a ResNet-18 backbone to further refine the \acrshort{ascb} output, yielding five feature sets $\{F_{2d}^{i}\}_{i=1}^{5}$ across different scales $i$, with each set $F_{2d}^{i} \in \mathbb{R}^{\frac{H}{2^{i}}\times \frac{W}{2^{i}} \times C_{i}}$. 
Suppose the radar point cloud comprising $N$ points, where $\textbf{r}_{2d}=\{(x_{2d}^{j}, y_{2d}^{j})\}_{j=1}^{N}$ represent $j^{th}$ point's projected pixel coordinate. 
Both these coordinates and the aforementioned multi-scale features are inputs to the point feature selection block.
\vspace{-4mm}
\paragraph{Adaptive sparse convolution block. }
The Sparse Convolutional Network \cite{sparsecnn} utilizes an observation mask in each sparse convolutional layer to filter out ``unobserved'' pixels from the input during the convolution. However, objects that are farther away from the ego-vehicle appear smaller on the image plane. This leads to a challenge that it may upsample the projected points into a wrong scale, since all points are treated equally by a single mask. 


Therefore, we propose the \acrshort{ascb}, which employs three binary observation masks to categorize radar detections by distance. Then, different convolution kernel sizes are selected for each group to enable precise feature propagation across different area sizes. 
Following a general statistical analysis of the projection size of common objects within our dataset, we select three distance range groups: $[0, 40)$, $[40, 70)$, and $[70, +\infty)$ meters. Within each range, a list of sparse convolutional layers with stride 1 is stacked to encode the input radar map according to the respective radar observation mask. We finally select the list of symmetric kernel sizes with $[11, 7, 7, 5, 5, 3]$, $[11, 7, 5, 5, 3, 3]$, and $[11, 7, 5, 3]$ for the aforementioned three distance groups. The outputs of these three ranges are element-wise summed to generate the final output of this block.
The detailed experiments are introduced in the supplementary material. 
\vspace{-4mm}
\paragraph{Point feature selection. }
After obtaining the multi-scale feature maps, we select the 2D features of each point at different scales. Thus, this block involves scaling the pixel coordinates according to the feature map's scale factor. At scale $i$, point features are extracted at the coordinates $(\lfloor \frac{x_{2d}^{j}}{2^{i}} \rfloor, \lfloor \frac{y_{2d}^{j}}{2^{i}} \rfloor)$, yielding a set of five point feature vectors $\{f_{2d}^{i}\}_{i=1}^{5}$, with $f_{2d}^{i}\in \mathbb{R}^{N\times C_{i}}$.
\begin{figure}[ht]
\centering
\vspace{-1mm}
\includegraphics[width=0.99\linewidth]{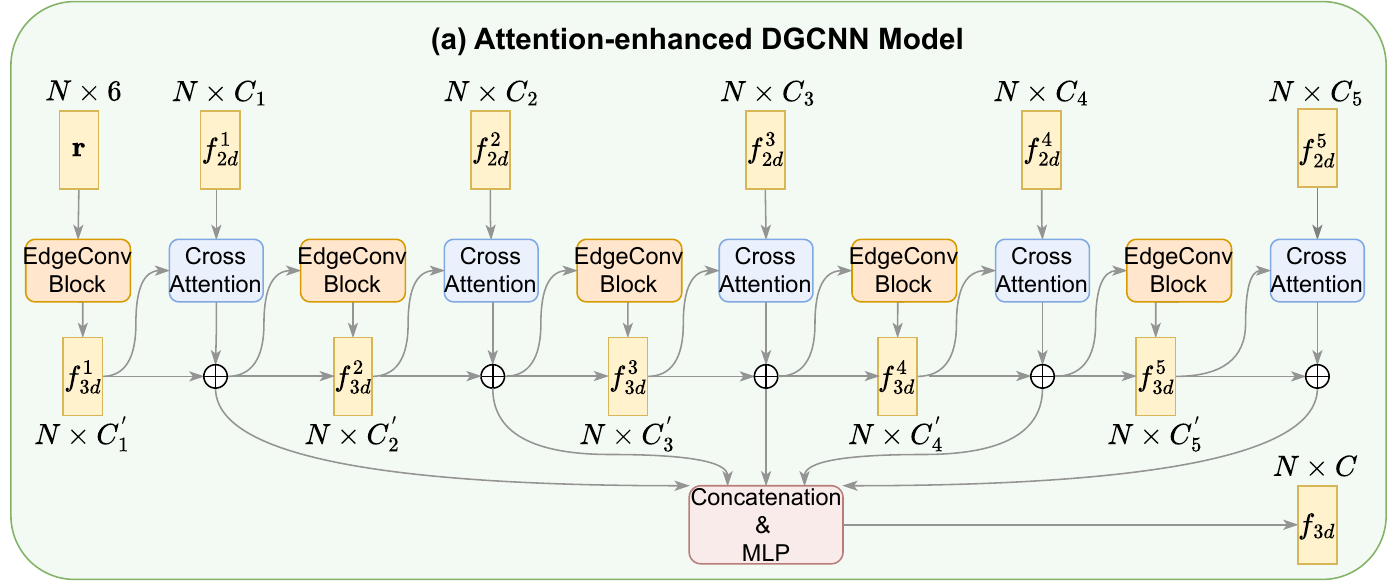}
\vspace{-2mm}
   \caption{Proposed attention-based DGCNN model, which incorporates extracted 2D features during the 3D feature generation, resulting in a robust representation of 3D radar features derived from sparse and noisy radar point clouds.}
\label{fig:DGCNN}
\vspace{-5mm}
\end{figure}

\vspace{-2mm}
\subsubsection{3D feature extraction submodule. }
\vspace{-1mm}

As illustrated in Figure \ref{fig:DGCNN}, our graph model effectively incorporates five EdgeConv blocks~\cite{dgcnn}, where the graph is constructed dynamically based on the \acrshort{knn} at each layer. This model aims to extract 3D point features from the input $\textbf{r}=\{(x_{3d}^{j}, y_{3d}^{j}, z_{3d}^{j}, v_{x}^{j}, v_{y}^{j}, rcs^{j})\}_{j=1}^{N}$, where the $j^{th}$ point is located at $(x_{3d}^{j}, y_{3d}^{j}, z_{3d}^{j})$ in 3D space.  The output of the first EdgeConv block $f_{3d}^{1}\in \mathbb{R}^{N\times C_{1}'}$, alongside $f_{2d}^{1}$, are input into a cross-attention block \cite{vaswani2017attention}, with $f_{3d}^{1}$ acting as the query and $f_{2d}^{1}$ generating the keys and values. 
We also add skip connection after the cross-attention to mitigate the potential gradient vanishing issue, resulting in the refined feature $f_{3d}^{1\prime}\in \mathbb{R}^{N\times C_{1}'}$
\vspace{-1mm}
\begin{equation}
\begin{aligned}
\resizebox{0.42\textwidth}{!}{$
    f_{3d}^{i\prime} = \text{Attention}(f_{3d}^{i}W_{3d_Q}^{i}, f_{2d}^{i}W_{3d_K}^{i}, f_{2d}^{i}W_{3d_V}^{i})+f_{3d}^{i}$}.
\end{aligned}
\vspace{-1mm}
\end{equation}

Subsequently, $f_{3d}^{1\prime}$ progresses to the next EdgeConv block. We repeat the EdgeCov block and cross-attention five times, yielding five intermediate features $\{f_{3d}^{i\prime}\}_{i=1}^{5}$, and they are further concatenated along the channel dimension processed by following MLP layers to generate the final feature output $f_{3d}\in \mathbb{R}^{N\times C}$.

\vspace{-2mm}
\subsubsection{2D-3D feature aggregation submodule. } 
\vspace{-1mm}
With $f_{3d}$ from the 3D feature extraction module and a set of 2D point features $\{f_{2d}^{i}\}_{i=1}^{5}$ at various scales $i$ from the 2D feature extraction module , the cross-attention operation uses $f_{3d}$ as the query and $f_{2d}^{i}$ as both key and value, yielding five aggregated features $\{f_{agg}^{i}\}_{i=1}^{5}$, each matching the dimensions of $f_{2d}^{i}$
\vspace{-1mm}
\begin{equation}
\resizebox{0.42\textwidth}{!}{$
    f_{agg}^{i} = \text{Attention}(f_{3d}W_{agg_Q}^{i}, f_{2d}^{i}W_{agg_K}^{i}, f_{2d}^{i}W_{agg_V}^{i})$}.
\vspace{-1mm}
\end{equation}

Furthermore, to derive a comprehensive global aggregated feature that spans all scales, the 2D features are concatenated across the channel dimension, and further apply cross-attention between this concatenated 2D feature and $f_{3d}$, resulting in a global aggregated feature $f_{agg}^{G} \in \mathbb{R}^{N\times C}$
\vspace{-1mm}
\begin{equation}
\begin{aligned}
    &f_{2d}^{G} = f_{2d}^{1} \otimes f_{2d}^{2} \otimes f_{2d}^{3} \otimes f_{2d}^{4} \otimes f_{2d}^{5}\\
    &f_{agg}^{G} = \text{Attention}(f_{3d}W_{agg_Q}^{G}, f_{2d}^{G}W_{agg_K}^{G}, f_{2d}^{G}W_{agg_V}^{G}),
\end{aligned}
\vspace{-1mm}
\end{equation}
where $\otimes$ signifies the concatenation of features along the channel dimension.


\vspace{-2mm}
\subsubsection{Point cloud upsampling submodule. }
\vspace{-1mm}
As a key module of GET-UP, we enhance radar data quality by leveraging the precision of LiDAR data, aiming to accurately identify and rectify the positioning of radar points. We start by detailing the approach for generating ground truth data, followed by the upsampling model architecture.
\vspace{-4mm}
\paragraph{Ground truth generation. }
Given the significant sparsity difference between LiDAR and radar point clouds—with radar detections being up to $1000\times$ sparser per frame \cite{nuscenes}—it is impractical to use the entire LiDAR dataset as ground truth for upsampling. Instead, a subset of $N_{L}$ LiDAR points is selected for ground truth supervision. A naive approach is randomly sampling these $N_{L}$ points from the LiDAR point cloud. However, this method fails to account for the spatial relevance of LiDAR points to actual radar detections. To address this, we refine our selection process by first calculating the Chamfer distance~\cite{chamfer} between radar and LiDAR points, then prioritize the $N_{L}$ LiDAR points with the smallest distances, effectively choosing those closest to the radar points as the upsampling ground truth $\textbf{R}_{gt}$.

\begin{figure}[ht]
\vspace{-1mm}
\centering
\includegraphics[width=0.99\linewidth]{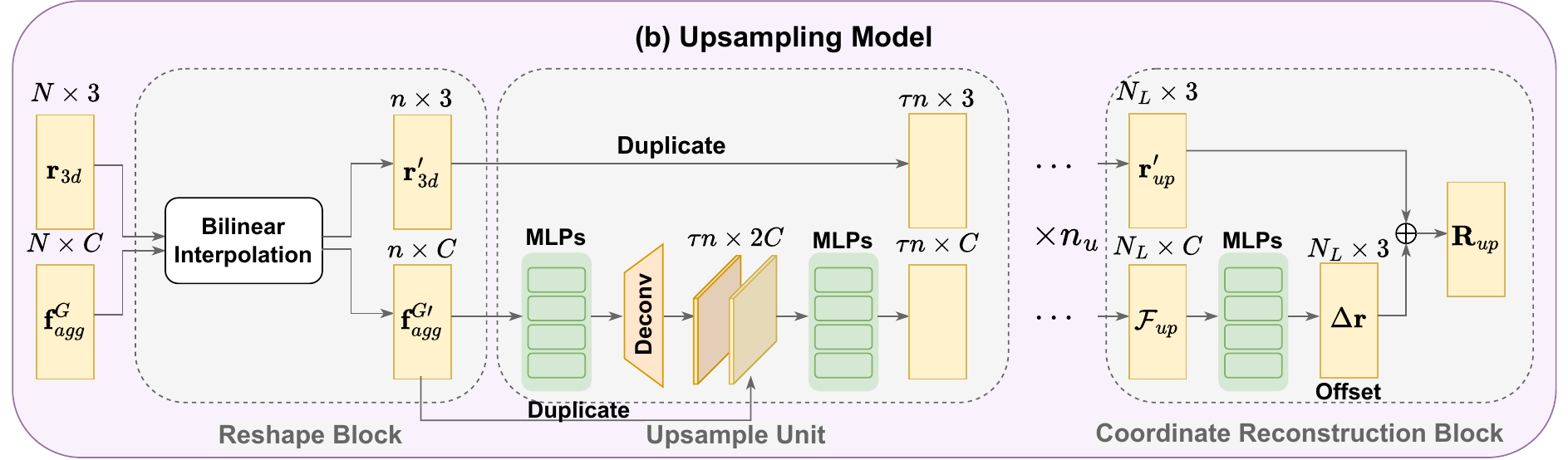}
\vspace{-2mm}
   \caption{Point cloud upsampling module. Initially, the 3D radar points and their associated features are processed by a reshape block, yielding a fixed number of points. Subsequently, they pass through \(n_{u}\) upsample units, each upsampling the inputs by a factor of \(\tau\). Ultimately, point offsets are derived from the processed features within the coordinate reconstruction block.}
\label{fig:upsampling}
\vspace{-4mm}
\end{figure}

\vspace{-4mm}
\paragraph{Upsampling model architecture. }Similar to existing methodologies~\cite{pugcn,pu_cascade}, our approach focuses on learning the offsets of target points rather than directly predicting their 3D positions. 
Nevertheless, unlike the typical point cloud upsampling task, we encounter the challenge that the quantity of radar points varies from frame to frame, resulting in an unpredictable upsampling ratio across different frames. To solve this problem, our upsampling model, visualized in Figure \ref{fig:upsampling}, contains three components: a \emph{reshape block}, $n_{u}$ \emph{upsample units} with the upsampling rate $\tau$, and a \emph{coordinate reconstruction block}. 

Initially, radar points $\textbf{r}_{3d}=\{(x_{3d}^{j}, y_{3d}^{j}, z_{3d}^{j})\}_{j=1}^{N}$, along with their global features $f_{agg}^{G}$, are processed by the reshape block, employing bilinear interpolation to scale the data to a predetermined number of points $n=\frac{N_{L}}{\tau^{n_{u}}}$. This generates a modified set of radar points $\textbf{r}_{3d}'=\{(x_{3d}^{j\prime}, y_{3d}^{j\prime}, z_{3d}^{j\prime})\}_{j=1}^{n}$ and their associated features $f_{agg}^{G\prime}\in \mathbb{R}^{n\times C}$.

Inspired by \cite{pu_cascade}, the process of our designed upsample unit involves duplicating the features $\tau$ times and concurrently processing them through a transposed convolutional layer to derive new point features. These resultant features are then concatenated along the channel axis and refined through two \acrshort{mlp} layers. Simultaneously, the input points are replicated by a factor of $\tau$.

Upon completing $n_{u}$ upsample units, we derive the upsampled feature vectors $\mathcal{F}_{up} = \{f_{up}^{j}\}_{j=1}^{N_{L}}$ and the duplicated points $\textbf{R}_{up}'=\{(X_{up}^{j\prime}, Y_{up}^{j\prime}, Z_{up}^{j\prime})\}_{j=1}^{N}$. The coordinate reconstruction block, utilizing $\mathcal{F}_{up}$ as input, computes per-point offsets $\Delta \textbf{r}$ through two \acrshort{mlp}s. These offsets are subsequently added to the duplicated points, resulting in the final upsampled 3D point cloud $\textbf{R}_{up}$.

This module returns two outputs, the upsampled point cloud $\textbf{R}_{up}$ and the upsampled features $\mathcal{F}_{up}$.

\vspace{-3.5mm}
\subsubsection{Feature refinement submodule. }
\vspace{-1mm}
This module augments the 2D features from two perspectives: utilizing aggregated features and incorporating upsampled features. 
Firstly, we enrich the 2D feature maps $\{F_{2d}^{i}\}_{i=1}^{5}$ by integrating the aggregated points features $\{f_{agg}^{i}\}_{i=1}^{5}$ at each respective scale $i$. More precisely, for the $j^{th}$ point at the $i^{th}$ scale, its feature is added back to the projected pixel coordinates $(\lfloor \frac{x_{2d}^{j}}{2^{i}} \rfloor, \lfloor \frac{y_{2d}^{j}}{2^{i}} \rfloor)$ on $F^{i}_{2d}$. In parallel, the global aggregated feature $f_{agg}^{G}$ and the upsampled feature $\mathcal{F}_{up}$ are projected onto the original, unscaled image plane served as a global feature map $F_{2d}^{0}$. 
Specifically, for the $j^{th}$ upsampled point $R^{j}_{up}=(X_{up}^{j}, Y_{up}^{j}, Z_{up}^{j})$, it is mapped to 2D coordinates using the camera's intrinsic and its associated upsampled feature is stored into $F_{2d}^{0}$. 
Importantly, any upsampled points in  $\textbf{R}_{up}$ that fall outside the original image plane after projection is discarded.

Finally, we concatenate $F_{2d}^{0}$ with the refined $\{F_{2d}^{i}\}_{i=1}^{5}$, yielding six comprehensive radar feature maps, which are passed to the decoder for the depth estimation task. Additionally, the upsampled 3D points $\textbf{R}_{up}$ are output to facilitate loss calculation for this specific branch.
\begin{figure}[ht]

\centering
\includegraphics[width=0.99\linewidth]{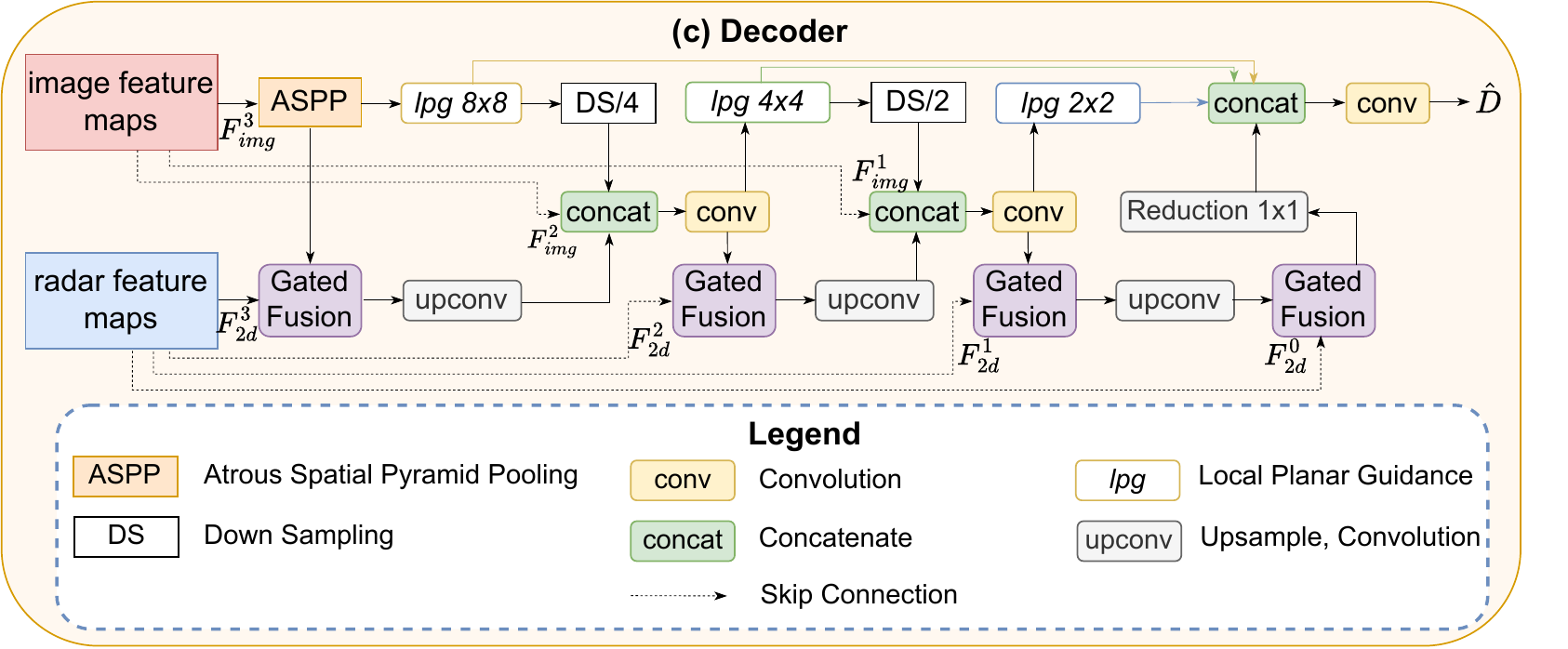}
\vspace{-2mm}
   \caption{Decoder architecture for depth estimation.}
\label{fig:bts_decoder}
\vspace{-5mm}
\end{figure}

\vspace{-1mm}

\begin{table*}[ht]
\vspace{-1mm}
\caption{Performance Comparison on nuScenes Official Test Set.}
\vspace{-2mm}
\label{tab:exp}
\resizebox{17.5cm}{!} 
{
\centering

\begin{tabular}{c||c|c|c|cccccccc}
\hline
\multirow{2}{*}{Eval Distance} & \multirow{2}{*}{Method} & \multicolumn{2}{c|}{Sensors} & \multicolumn{8}{c}{Metrics} \\ \cline{3-12} 
                                    &                         & Image      & Radar      & MAE $\downarrow$ & RMSE $\downarrow$ & AbsRel $\downarrow$  &log10 $\downarrow$ & RMSElog $\downarrow$ & $\delta_{1}$ $\uparrow$& $\delta_{2}$ $\uparrow$ & $\delta_{3}$ $\uparrow$  \\ \hline
\multirow{8}{*}{50m}                & BTS \cite{bts}  & \cmark &            & 1.937 & 3.885 & 0.116                                            &0.045 & 0.179 & 0.883 & 0.957 & 1.937 \\ 
                                    & RC-PDA \cite{rc-pda}   & \cmark & \cmark & 2.225 & 4.159 &  0.106   & 0.051  & 0.186 & 0.864 & 0.944 & 0.974      \\ 
                                    & RC-PDA-HG \cite{rc-pda}  & \cmark & \cmark & 2.210 & 4.234 & 0.121   & 0.052  & 0.194  & 0.850  & 0.942  & 0.975       \\ 
                                    & DORN  \cite{dorn_radar}  & \cmark & \cmark & 1.898 & 3.928 & 0.100 &   0.050 & 0.164 & 0.905 & 0.962 & 0.982  \\ 
                                    & RadarNet   \cite{radarnet}  & \cmark &  \cmark & 1.706 & 3.742 & 0.103 & 0.041 & 0.170 & 0.903 & 0.965 &0.983 \\ 
                                    & CaFNet \cite{sun2024cafnet}   & \cmark &  \cmark & 1.674 & 3.674 & 0.098 & 0.038 & 0.164 & 0.906 & 0.963 & 0.983 \\
                                    & Li \textit{et al.} \cite{li2023sparse} & \cmark &  \cmark & 1.524 & 3.567 & - & - & - & - & - & - \\
                                    & GET-UP (Ours)      & \cmark &  \cmark  & \textbf{1.241} & \textbf{2.857} & \textbf{0.072} & \textbf{0.030} & \textbf{0.135} & \textbf{0.943}  & \textbf{0.977} & \textbf{0.988}\\
                                    \hline
\multirow{8}{*}{70m}                & BTS \cite{bts} & \cmark &            & 2.346 & 4.811& 0.119 & 0.047                                     & 0.188 & 0.872 & 0.952 & 0.979       \\ 
                                    & RC-PDA  \cite{rc-pda}   & \cmark & \cmark & 3.338 & 6.653 & 0.122  & 0.060 & 0.225 & 0.822 & 0.923 & 0.965  \\ 
                                    & RC-PDA-HG  \cite{rc-pda}  & \cmark & \cmark & 3.514 & 7.070 & 0.127  & 0.062 & 0.235 & 0.812 & 0.914 & 0.960      \\ 
                                    & DORN  \cite{dorn_radar}  & \cmark & \cmark & 2.170 & 4.532 & 0.105  & 0.055 & 0.170 & 0.896 & 0.960 & 0.980       \\ 
                                    & RadarNet    \cite{radarnet}  & \cmark &  \cmark & 2.073 & 4.591 & 0.105  & 0.043 & 0.181 & 0.896 & 0.962 & 0.981      \\ 
                                    & CaFNet \cite{sun2024cafnet}& \cmark &  \cmark &  2.010 & 4.493 & 0.101 & 0.040 & 0.174 & 0.897 & 0.961  & 0.983 \\
                                    & Li \textit{et al.} \cite{li2023sparse} & \cmark &  \cmark & 1.823 & 4.304 & - & - & - & - & - & - \\
                                    
                                    &  GET-UP (Ours) & \cmark &  \cmark &  \textbf{1.541} & \textbf{3.657} & \textbf{0.075} & \textbf{0.032} & \textbf{0.145} & \textbf{0.936} & \textbf{0.974} & \textbf{0.986} \\
                                    \hline
\multirow{13}{*}{80m}                & BTS \cite{bts}   & \cmark &            & 2.467 & 5.125 & 0.120 & 0.048 & 0.191 & 0.869 & 0.951 & 0.979 \\ 
                                    & AdaBins \cite{adabins} & \cmark &            & 3.541 & 5.885 & 0.197& 0.089 & 0.261 & 0.642 & 0.929 & 0.977       \\ 
                                    & P3Depth \cite{p3depth}  & \cmark &            & 3.130 & 5.838 & 0.165& 0.065 & 0.222 & 0.804 & 0.934 & 0.974        \\ 
                                    & LapDepth \cite{lapdepth}   & \cmark &            & 2.544 & 5.151 & 0.117 & 0.049 & 0.187 & 0.865 & 0.953 & 0.980       \\ 
                                    & S2D$^{\dag}$ \cite{lidar_2d1}  & \cmark & \cmark  & 2.374  & 5.628   & 0.115  & -   & -  & 0.876  & 0.949  & 0.974  \\
                                    & RC-PDA    \cite{rc-pda}  & \cmark & \cmark & 3.721 & 7.632  &  0.126 & 0.063 & 0.238 & 0.813 & 0.914 & 0.960       \\ 
                                    & RC-PDA-HG  \cite{rc-pda}  & \cmark & \cmark & 3.664 & 7.775 & 0.138 & 0.064 & 0.245 & 0.806 & 0.909 & 0.957       \\ 
                                    & DORN  \cite{dorn_radar}  & \cmark & \cmark & 2.432 & 5.304 & 0.107 & 0.056 & 0.177 & 0.890 & 0.960 & 0.981      \\ 
                                    & RCDPT$^{\dag}$ \cite{rcdpt}  & \cmark & \cmark & - & 5.165 & 0.095 &  - & - &0.901 & 0.961 & 0.981      \\ 
                                    & RadarNet  \cite{radarnet}  & \cmark & \cmark & 2.179 & 4.899 & 0.106 & 0.044 & 0.184 & 0.894 & 0.959 & 0.980 \\ 
                                    & CaFNet \cite{sun2024cafnet}  & \cmark & \cmark & 2.109 & 4.765 & 0.101 & 0.040 & 0.176 & 0.895 & 0.959 & 0.981\\
                                    & Li \textit{et al.} \cite{li2023sparse} & \cmark &  \cmark & 1.927 & 4.610 & - & - & - & - & - & - \\
                                    & GET-UP (Ours)  & \cmark & \cmark & \textbf{1.632} & \textbf{3.932} & \textbf{0.076} & \textbf{0.032} & \textbf{0.148} & \textbf{0.934} & \textbf{0.974} & \textbf{0.986}\\
                                    \hline
\end{tabular}
}
\begin{tablenotes}
      \small
      \item $^{\dag}$ These results come from the paper that tests the model performance on a different test set. This leads to the metrics being less comparable.
    \end{tablenotes}
     \vspace{-4mm}
\end{table*}

\subsection{Decoder}
\label{subsec:decoder}
\vspace{-1mm}
Our depth estimation framework is built upon the BTS model \cite{bts}, leveraging the local planar guidance concept to enhance the upsampling process and extract more meaningful features. In this process, we incorporate both image features $\{F_{img}^{i}\}_{i=1}^{5}$ and radar features $\{F_{2d}^{i}\}_{i=0}^{5}$ as input and employ the gated fusion technique \cite{radarnet} to fuse radar and image features. Detailed visualizations of the decoder's architecture are provided in Fig. \ref{fig:bts_decoder}.
\vspace{-1mm}

\subsection{Loss Functions}
\vspace{-1mm}
Our model employs two loss functions to facilitate depth estimation and point cloud upsampling tasks. 
To guide the depth estimation task, we follow methodologies from \cite{radarnet}, accumulating LiDAR point clouds from neighboring frames to construct an accumulated depth map. Subsequently, we apply the scaffolding technique \cite{scrffolding} to generate a dense depth map $D$. As demonstrated in \cite{li2023sparse}, supervised by single scan depth $D_{s}$ improves the depth prediction accuracy. Thus, we utilize $D_{s}$ and $D$ to supervise our depth estimation task as follows: 
\begin{equation}
\resizebox{0.48\textwidth}{!}{$
    L_{Depth} = \frac{1}{|\Omega_{s}|}\sum_{x\in \Omega_{s}}|D_{s}(x) - \hat{D}(x)| + \frac{1}{|\Omega|}\sum_{x\in \Omega}|D(x) - \hat{D}(x)|,$}
\end{equation}
which only calculated within the sets of pixels where $D_{s}$ or $D$ are valid.

The Chamfer distance loss \cite{chamfer} is utilized to reduce the discrepancy between the upsampled point cloud $\textbf{R}_{up}$ and the ground truth $\textbf{R}_{gt}$:
\begin{equation}
\resizebox{0.48\textwidth}{!}{$
    L_{Up} = \frac{1}{|\textbf{R}_{up}|}\sum _{p\in \textbf{R}_{up}}min_{q\in \textbf{R}_{gt}} ||p-q||_{2}^{2} + 
    \frac{1}{|\textbf{R}_{gt}|}\sum _{q\in \textbf{R}_{gt}}min_{p\in \textbf{R}_{up}} ||p-q||_{2}^{2}$.}
\end{equation}
Here, $p$ represents a 3D point in $\textbf{R}_{up}$, and $q$ denotes a 3D point in $\textbf{R}_{gt}$. The term $||\cdot||_{2}^{2}$ signifies the squared Euclidean distance.

The final loss function is a weighted sum of the individual losses:
$L=L_{Depth} + \alpha L_{Up}$, where $\alpha$ is a weighting factor to balance the importance of the two tasks.


\section{Experiments}
\label{sec:experiments}
\vspace{-1mm}
This section first introduces the dataset and the implementation details. Then, we describe the evaluation metrics and compare our GET-UP with the existing methodologies in the quantitative and qualitative aspects. Finally, we conduct ablation studies to further underscore our proposed methods' effectiveness.
\begin{figure*}[t]
    \centering
    \includegraphics[width=0.95\textwidth]{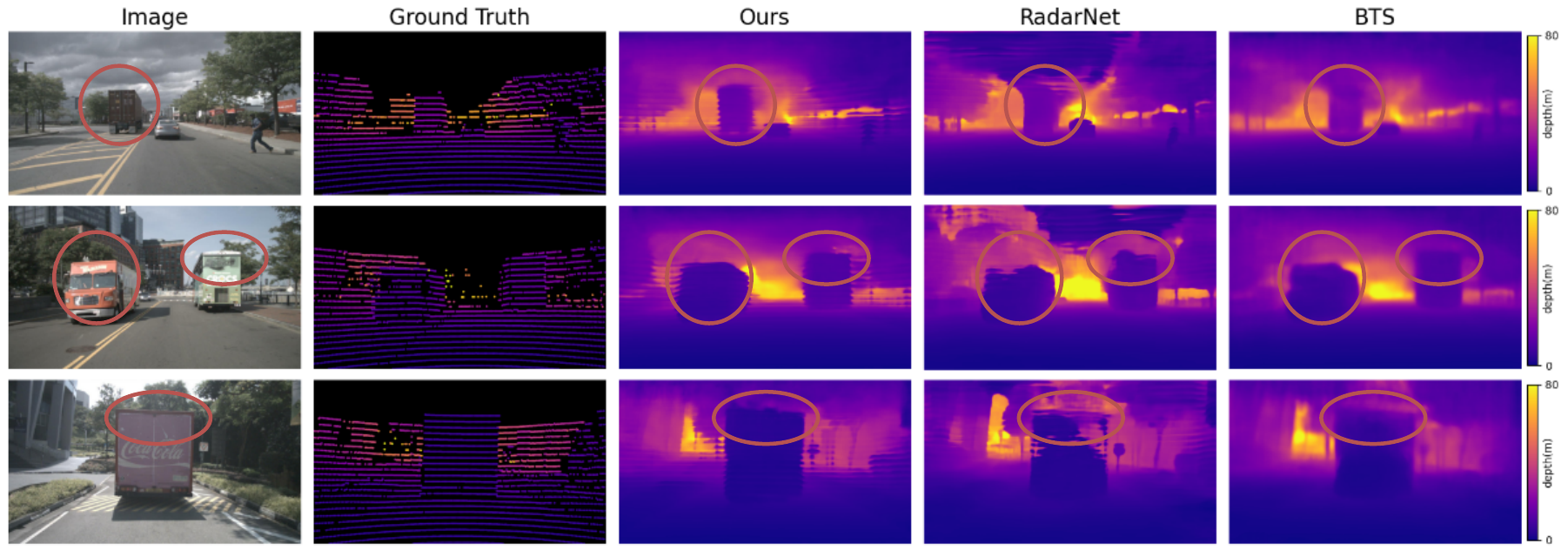}
    
       \caption{Qualitative comparison on nuScenes test set. Column 1 shows the RGB image; column 2 plots the ground truth depth map. We compare our result with the RadarNet and our baseline BTS at 80 meters depth range.}
    \label{fig:qualitative}
    \vspace{-4mm}
    \end{figure*}
\subsection{Dataset and Implementation Details}
We utilize the nuScenes dataset \cite{nuscenes}, a comprehensive multi-sensor dataset dedicated to autonomous driving research, for the training and evaluation of our model. The dataset is partitioned into training, validation, and test subsets, consisting of 700, 150, and 150 scenes, respectively. We utilize the front camera and radar data to train and evaluate our model. Notably, we employ a single radar scan as input and our model can handle undefined number of radar points as input. 

For training, we aggregate 80 preceding and 80 succeeding LiDAR frames to create an accumulated depth map and then employ the scaffolding technique \cite{scrffolding} to generate a dense depth map $D$. It's important to note that for evaluation, the single-frame LiDAR depth map $D_{gt}$ serves as the ground truth. In the point cloud upsampling task, we sample 128 LiDAR points per frame as the ground truth $\textbf{R}_{gt}$, normalizing these points and the radar points using their centroid and maximum distance. 
The upsampled points estimated by the feature refinement submodule are then de-normalized
before projection onto the image plane.

Our model is developed in PyTorch \cite{Pytorch} and trained on an Nvidia\textsuperscript{\textregistered} Tesla A30 GPU with a batch size of 6. We employ the Adam optimizer \cite{kingma2017adam} with an initial learning rate of $1e^{-4}$, adjusting it according to a polynomial decay rate with a power of $p=0.9$. To prevent overfitting, image augmentation techniques such as random flips and adjustments to contrast, brightness, and color are applied. Moreover, random cropping to a size of $352 \times 704$ pixels is conducted during training to enhance the model robustness further.
\vspace{-1mm}
\subsection{Quantitative Results}
In this study, we benchmark our GET-UP model against both image-based methods \cite{bts,adabins,lapdepth,p3depth} and existing radar-camera depth estimation approaches \cite{rc-pda,rcdpt,radarnet,dorn_radar,sun2024cafnet,li2023sparse}, using the standard evaluation metrics, detailed in the supplementary material. The models are assessed on the official nuScenes test set across three evaluation ranges: up to 50 meters, 70 meters, and 80 meters, with detailed results presented in Table \ref{tab:exp}.

Our GET-UP model demonstrates superior performance over image-only methods across all metrics. Specifically, it enhances the BTS baseline \cite{bts} by $33.8\%$ in MAE and $23.3\%$ in RMSE, highlighting the substantial benefits of incorporating radar data into image-based methods. Furthermore, we explored LiDAR-camera depth completion techniques \cite{lidar_2d1,lidar_compare2} using radar data to generate sparse depth maps. These attempts yielded unsatisfactory outcomes due to the even sparser and more ambiguous nature of radar-generated depth maps. Techniques such as those in \cite{PointFusion,fuseNet,PointDC} prove inappropriate for radar-camera depth estimation tasks because they rely on a predetermined number of LiDAR points as inputs, which is incapable of handling various number of radar points.

Following the methodology of \cite{radarnet}, our approach utilizes a single radar scan, contrasting with methods like \cite{dorn_radar,rc-pda,rcdpt} that employ multiple radar scans to enhance point cloud density. Remarkably, our method delivers superior results with fewer radar points compared to these approaches. Remarkably, GET-UP outperforms \cite{li2023sparse} by $18.6\%$, $18.3\%$, and $15.3\%$ in MAE and $19.9\%$, $15.0\%$, and $14.7\%$ in RMSE at the 50, 70, and 80-meter evaluation distances, respectively, demonstrating its efficacy in leveraging limited radar data for accurate depth estimation.

\vspace{-1mm}

    

\vspace{-1mm}
\subsection{Qualitative Results}
Fig. \ref{fig:qualitative} showcases a comparative analysis of our GET-UP method against the baseline \cite{bts} and RadarNet \cite{radarnet}. 
Overall, our proposed GET-UP predicts depth maps with clearer object boundaries compared to RadarNet and the baseline, both at long and short ranges. For example, in the first row, GET-UP effectively distinguishes between the sky and the upper boundary of the track at a far distance. In the second and third rows, our method demonstrates greater robustness by accurately predicting the shapes of various objects.


\subsection{Ablation Study}
To further ascertain the efficiency of our GET-UP, we conduct a series of ablation studies to verify the effectiveness of each component. First, we analyze the impact of the \acrshort{ascb}. Secondly, the efficacy of the 3D feature extraction submodule is evaluated. At last, we quantify the reliability of the point cloud upsampling submodule. 
\vspace{-3mm}
\subsubsection{Adaptive sparse convolution block. }
\vspace{-1mm}

Initially, we conducted experiments without utilizing any sparse convolution refinement, which directly extracts 2D features from radar projections. Subsequently, we compared our proposed \acrshort{ascb} against the conventional sparse convolution block~\cite{sparsecnn}, which employs a single mask alongside a sequence of kernel sizes set at $[11, 7, 5, 3, 3]$. Compared to the conventional sparse convolution block, our \acrshort{ascb} improves the 
RMSE with $7.7\%$. 
\setlength{\tabcolsep}{7pt}


\begin{table}[ht]
\vspace{-3mm}
\centering
\caption{Ablation study on the sparse convolution block.}
\vspace{-2mm}
\resizebox{8cm}{!} 
{
\begin{tabular}{cc||cccc}
\hline
 conventional~\cite{sparsecnn} & \acrshort{ascb} & MAE $\downarrow$ & RMSE $\downarrow$ & AbsRel $\downarrow$ & $\delta_{1}$ $\uparrow$\\ \hline
\xmark &  \xmark & 1.852 & 4.432 & 0.093 & 0.909 \\
\cmark & \xmark & 1.792 & 4.262 & 0.088 & 0.918 \\
\xmark & \cmark & \textbf{1.632} & \textbf{3.932} & \textbf{0.076} & \textbf{0.934} \\ 
 \hline
\end{tabular}
}
\label{table:ablation_scm}
\vspace{-6mm}
\end{table}
\vspace{-1mm}
\subsubsection{Radar 3D feature extraction submodule. }
\vspace{-1mm}

In this section, we first conduct an experiment without 3D feature extraction. Thus, $x_{Radar}$ is processed solely through the \acrshort{ascb} and then by the ResNet encoder. The result underscores the effectiveness of incorporating geometry information and the critical role of the 2D-3D feature aggregation and refinement process in enhancing model performance.
Then, we benchmark our attention-enhanced DGCNN against established models such as GCN \cite{gcn}, GCN2 \cite{gcn2}, and the original DGCNN \cite{dgcnn} architectures. Furthermore, we explore the optimal number of nearest neighbors ($k$) for each radar point to determine the most effective value, with the comparative results presented in Table \ref{table:ablation_gnn}. The findings clearly demonstrate the superior performance of our attention-enhanced model. Notably, a larger $k$ value degrades performance since it tends to only capture global features due to the sparse nature of radar points. 
This leads to errors during feature extraction, underscoring the importance of carefully selecting $k$ to balance detail capture and noise minimization. Our proposed 3D feature extraction module improves the MAE by $6.5\%$ compared to the solely 2D feature extraction architecture.
\setlength{\tabcolsep}{7pt}

\begin{table}[ht]
\caption{Ablation study on the GNN models.}
\vspace{-2mm}
\resizebox{8.2cm}{!} 
{
\centering
\begin{tabular}{cc||cccc}
\hline
Model & k-value & MAE $\downarrow$ & RMSE $\downarrow$ & REL $\downarrow$ & $\delta_{1}$ $\uparrow$\\ \hline
w/o GNN & N/A & 1.884 & 4.663 & 0.090 & 0.911 \\
GCN & 3  & 1.877 & 4.432 & 0.087 & 0.918 \\
GCN & 4  & \textbf{1.852} &\textbf{4.401} & \textbf{0.085} & \textbf{0.921} \\
GCN & 6  & 1.882 & 4.441 & 0.088 & 0.917 \\
GCN & 8  & 1.894 & 4.485 & 0.087 & 0.918 \\
GCN & 10  & 1.896 & 4.489 & 0.089 & 0.916 \\
\hline\hline
GCN & 4  & 1.852 & 4.401 & 0.085 & 0.921 \\
GCN2 & 4  & 1.874 & 4.520 & 0.088 & 0.919 \\
DGCNN & 4  & 1.843 & 4.448 & 0.083 & 0.922 \\
Ours &  4 & \textbf{1.762} & \textbf{4.331} & \textbf{0.081} & \textbf{0.925} \\ 
 \hline
\end{tabular}
}
\label{table:ablation_gnn}
\vspace{-4mm}
\end{table}

\vspace{-2mm}
\subsubsection{Point cloud upsampling submodule. }
\vspace{-1mm}
To demonstrate the effectiveness of this module, we initially perform an experiment excluding the upsampling task. Subsequently, we conduct further experiments to identify the optimal number of upsampling units $n_{u}$ and the number of upsampled points $N_{L}$. All other components of the model remain unchanged during these evaluations. 
The results demonstrate that incorporating point cloud upsampling as an auxiliary task substantially enhances depth estimation accuracy. Moreover, given that the average count of radar points per frame is approximately 60, selecting an appropriate value for $N_{L}$ is crucial to ensure the efficacy of the upsampling process. 



\setlength{\tabcolsep}{7pt}
\begin{table}[ht]
\vspace{-1mm}
\caption{Ablation study on the upsampling module.}
\vspace{-2mm}
\resizebox{8.2cm}{!} 
{
\centering
\begin{tabular}{ccc||cccc}
\hline
Upsampling & $n_{u}$ \quad & $N_{L}$ \quad& MAE $\downarrow$ & RMSE $\downarrow$ & REL $\downarrow$ & $\delta_{1}$ $\uparrow$\\ \hline

\xmark & N/A & N/A  & 1.762 & 4.331 & 0.081 & 0.925 \\

\cmark & 1  & 128 & 1.721 & 4.231 & 0.079 &0.928 \\
\cmark & 2  & 128 &  \textbf{1.632} & \textbf{3.932} & \textbf{0.076} & \textbf{0.934}  \\
\cmark & 3   & 128 & 1.679 & 3.973 & 0.078 & 0.931 \\
\hline\hline

\cmark & 2 & 64 & 1.702 & 4.090 & 0.080 & 0.929 \\
\cmark & 2 & 128 &  \textbf{1.632} & \textbf{3.932} & \textbf{0.076} & \textbf{0.934}  \\
\cmark & 2 & 256  & 1.683 & 3.969 & 0.077 & 0.932 \\
 \hline
\end{tabular}
}
\label{table:ablation_pu}
\vspace{-5mm}
\end{table}



\section{Conclusion}
\label{sec:conclusion}

In this paper, we propose GET-UP, a geometry-aware algorithm designed to tackle the significant challenges in radar-camera depth estimation due to the inherent ambiguity and sparsity of radar data.
Our approach integrates both 2D and 3D representations of radar data, utilizing an attention-enhanced DGCNN model for the extraction of 3D features without compromising 2D spatial context. To address the issue of radar data sparsity, we implement two strategies: the \acrshort{ascb}, which densifies radar data on the 2D plane to facilitate the extraction of 2D features and a point cloud upsampling task that enhances radar point density from a 3D perspective. GET-UP sets a new benchmark on the nuScenes dataset, improving $15.3\%$ in MAE and $14.7\%$ in RMSE over the previously best-performing model. Looking ahead, exploring diverse upsampling algorithms on radar point clouds and refining the integration of 2D and 3D radar features present valuable directions for further research.


\section{Acknowledgement}
Research leading to these results 
has received funding from the EU ECSEL Joint Undertaking under grant agreement n° 101007326 (project AI4CSM) and from the partner national funding authorities the German Ministry of Education and Research (BMBF).
\label{sec:acknowledgement}

{\small
\bibliographystyle{ieee_fullname}
\bibliography{egbib}
}

\end{document}